\documentclass[letterpaper, 10 pt, conference]{ieeeconf}  % Comment this line out if you need a4paper

\IEEEoverridecommandlockouts                              % This command is only needed if
\usepackage{amsmath} % assumes amsmath package installed
\usepackage{graphics} % for pdf, bitmapped graphics files
\usepackage{epstopdf}
\usepackage{epsfig}
\usepackage{color}
\usepackage[T1]{fontenc}
\usepackage{multirow}
\title{\LARGE \bf
Compliant Movement Primitives in a Bimanual Setting
}

\author{Aleksandar Batinica$^{1}$, Bojan Nemec$^{2}$, Ale\v{s} Ude$^{2}$,  Mirko Rakovi\'{c}$^{1}$ and Andrej Gams$^{2}$% <-this % stops a space
\thanks{*This work was supported by the Slovenian Research Agency project ARRS BI-RS$/$16-17-048.}% <-this % stops a space
\thanks{$^{1}$Aleksandar Batinica and Mirko Rakovi\'{c} are with the Faculty of Technical Sciences, University of Novi Sad, Novi Sad, Serbia.
        {\tt\small batinica@uns.ac.rs}}%
\thanks{$^{2}$Bojan Nemec, Ale\v{s} Ude and Andrej Gams are with Humanoid and Cognitive Robotics Lab, Dept. of Automatics, Biocybernetics and Robotics, Jo\v{z}ef Stefan Institute, Ljubljana, Slovenia.
        {\tt\small name.surname@ijs.si}}%
}

\begin{document}

\maketitle
\thispagestyle{empty}
\pagestyle{empty}

%%%%%%%%%%%%%%%%%%%%%%%%%%%%%%%%%%%%%%%%%%%%%%%%%%%%%%%%%%%%%%%%%%%%%%%%%%%%%%%%
\begin{abstract}

Simultaneously achieving low trajectory errors and compliant control \emph{without} explicit models of the task was effectively addressed with Compliant Movement Primitives (CMP). For a single-robot task, this means that it is accurately following its trajectory, but also exhibits compliant behavior in case of perturbations. In this paper we extend this kind of behavior without explicit models to bimanual tasks. In the presence of an external perturbation on any of the robots, they will both move in synchrony in order to maintain their relative posture, and thus not exert force on the object they are carrying. Thus, they will act compliantly in their absolute task, but remain stiff in their relative task. To achieve compliant absolute behavior and stiff relative behavior, we combine joint-space CMPs with the well known symmetric control approach. To reduce the necessary feedback reaction of symmetric control, we further augment it with copying of a virtual force vector at the end-effector, calculated through the measured external joint torques. Real-world results on two Kuka LWR-4 robots in a bimanual setting confirm the applicability of the approach.
\end{abstract}

%%%%%%%%%%%%%%%%%%%%%%%%%%%%%%%%%%%%%%%%%%%%%%%%%%%%%%%%%%%%%%%%%%%%%%%%%%%%%%%%
\section{INTRODUCTION} \label{sec:intro}
Until recently robots have been exclusive to industrial environments. Due to high stiffness and position control to accomplish accurate execution of their given tasks, robots were deemed hazardous to humans and unexpected objects in their workspace and therefore confined to cages \cite{Haddadin2014}. The development of robotics has led to the notion of collaborative robotics, where both the human and robot share their workspace to accomplish a task \cite{Haddadin2009}. Collaborative robotics spans beyond the factory work-floor to everyday human environments, such as household, hospitals, etc., and includes also bimanual and humanoid robots.

In shared workspaces safety of the human is of primary concern. One approach to ensuring safety is through the compliance of the robot. This can be ensured through contact detection, for example using an artificial tactile skin \cite{Mittendorfer11}. Passive compliance can be achieved with elastic elements, which can even be actively adapted, for example with variable stiffness actuators \cite{bicchi2004}. Compliance can also be achieved by implementing appropriate active torque strategies, which rely on comparing the actual torques and the required theoretical torques \cite{hogan1987}. However, this requires the correct dynamic model of not only the robot, but also of each task and task variation. Models of the task dynamics are often not available or hard to obtain \cite{Petric2010a}.

One way of mitigating the need to develop dynamical models of tasks is to learn or record the specific required torques for the given task with learning by imitation. These torques are then applied for the repetition of the exact same task. The framework of Compliant Movement primitives (CMPs) \cite{Denisa2016}, applicable to robots with active torque control, utilizes this approach. The method was extended to generalize between a set of learned situations in order to generate the torques for a new task variation, such as a different load or speed. Thus, a single robot was able to perform a wide variety of task variations through direct joint-position and joint-torque control, with low trajectory errors but compliantly in case of an external perturbation. In this paper we extend the original CMP approach to bimanual tasks.
	\begin{figure}[t]
    \vspace{2mm}
	\centering
	\includegraphics[width = 0.5\textwidth]{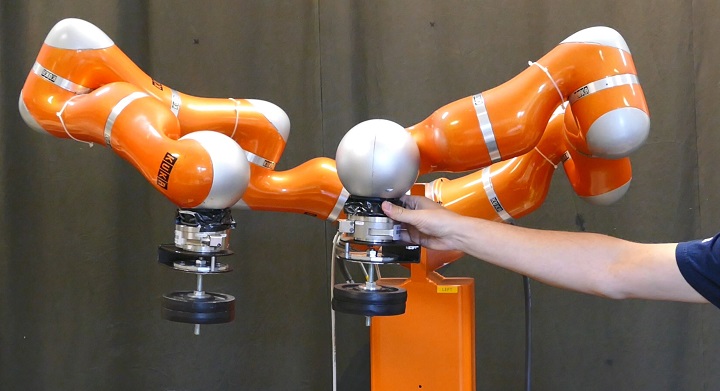}
	\caption{A person interacting with a compliant bimanual robot setup, where the bimanual task is preserved.}
	\label{fig:intro}\vspace{-6mm}
	\end{figure}

\subsection{Problem Statement}\label{sec:statement}
In this paper we investigate compliant control of a bimanual robotic system \emph{without} explicit dynamical models of the task. Therefore the control of the bimanual system must enable:
\begin{itemize}
  \item synchronous bimanual behavior with high relative robot-robot position error rejection,
  \item compliant behavior of the bimanual system in case of external perturbations, and
  \item low trajectory tracking errors when there are no perturbations.
  \item This must be achieved without explicit dynamical models, but with the use of task-specific torques within the framework of CMPs.
\end{itemize}
%In the paper we also show that the proposed method can be generalized to new task parameters.

\subsection{Related Work}
Related work can be separated into two distinct topics: compliant control and bimanual tasks.

\emph{Compliant control} typically relies on explicit dynamics of the robot and the task \cite{Peters2011}. However, besides the above-mentioned CMPs, which are at the core of this paper, similar approaches that rely on task-specific models have emerged. For example in \cite{Calandra2015} tactile sensors were used to determine the force of contact with the environment on the iCub robot. This information was used calculate the joint-torques from the measured arm pose, and use them in a feed-forward manner in control. Similarly, joint torques along the kinematic trajectory were encoded as DMPs and used as the feed-forward signal to increase the accuracy in the next execution of the in-contact task in \cite{Steinmetz2015}.

\emph{Bimanual control} of robots can be either asymmetric or symmetric. While the former controls each robot independently, the latter considers both robots as a single system. An example of an asymmetric control scheme using motion primitives is described in \cite{Gams14a}. There the robots are coupled through feed-forward signals learned in a few iterations from force-feedback. However, the robots are stiff. A system that combines bimanual robot operation based on dynamical systems was presented in \cite{Salehian-RSS-16}. In the paper the motion of robotic arms is adapted for coordinated bimanual receiving/intercepting an object. The system also relies on a virtual object to generate the necessary motions.

On the other hand, a symmetric system can fully characterize a cooperative operational space and allow the user to specify the task in relative and absolute coordinates, resulting in geometrically meaningful motion variables defined at the position/orientation level \cite{Chiacchio98}. An example of such is \cite{Nemec2016}, where a human-robot cooperation scheme
for bimanual robots was presented. Similar to our proposed algorithm, \cite{Nemec2016} relies on separately defining the gains for absolute and relative motion, however it does not allow for low trajectory tracking errors when the absolute gains are set low. Other physical human-robot interaction schemes were investigated in the past, for example in \cite{Adorno2010, Mortl2012, Park2015}.

%\emph{Generalization} has been extensively applied in robotics, with methods such as Locally Weighted Regression \cite{atkeson1997} and Gaussian Process Regression (GPR) \cite{rasmussen2006} at the forefront. We refer the reader to \cite{Calinon2015} for an extensive overview of generalization of kinematic behavior. However, not so many approaches deal with dynamic variables. An example of generalization of force trajectories for peg-in-hole operation is given in \cite{Kramberger2016}. In our paper we rely on GPR to generate new CMPs, similar to \cite{Denisa2016}, but for a bimanual system.

In the next Section we introduce the CMP framework. Section \ref{sec:Bimanual} gives the basics on the bimanual kinematics explains the complete controller.
%Generalization in introduced in Section \ref{sec:generalization}.
Applicability of the approach is shown through results in Section \ref{sec:results}. Various aspects of the approach are discussed in Section \ref{sec:discussion}.

\section{COMPLIANT MOVEMENT PRIMITIVES} \label{sec:CMPs}
\subsection{Robot Control}
The controller of an impedance-controlled robot, such as the Kuka LWR-4 robot \cite{Bischoff2010}, is defined by
	\begin{equation}
	\label{eq:kuka_ctrl}
	\vec{\tau}_u = \textbf{K}_q(\vec{q}_d - \vec{q}\,) + \textbf{D}_q(\dot{\vec{q}}_d - \dot{\vec{q}\,}) + \vec{f}_{dynamic}(\vec{q}, \dot{\vec{q}}, \ddot{\vec{q}}\,),
	\end{equation}
where $\vec{\tau}_u$ is the control torque vector sent to the actuators, $\textbf{K}_{q}$ is a diagonal joint-stiffness matrix, $\vec{q}_d$ and $\vec{q}$ are the vectors of the desired and measured joint positions, respectively, $\textbf{D}_q$ is a diagonal damping matrix, $(\vec{\dot{q}}_d$ and $\vec{\dot{q}}$ are the desired and measured vectors of joint velocities, respectively, and $\vec{f}_{dynamic}(\vec{q}, \dot{\vec{q}}, \ddot{\vec{q}})$ represents the robot dynamics and all the non-linearities occurring in the robot (Coriolis, friction, ...)

The robot can be made compliant by lowering the stiffness ($\textbf{K}_q$), however, this also lowers the trajectory tracking capability of the robot. To compensate, feed-forward torques $\vec{\tau}_{ff}$ are added to the motor torque to preserve trajectory tracking, resulting in
\begin{equation}
	\label{eq:kuka_ctrl2}
	\vec{\tau}_u = \textbf{K}_q(\vec{q}_d - \vec{q}\,) + \textbf{D}_q(\dot{\vec{q}}_d - \dot{\vec{q}\,}) + \vec{f}_{dynamic}(\vec{q}, \dot{\vec{q}}, \ddot{\vec{q}}\,) + \vec{\tau}_{ff}.
	\end{equation}
Typically, feed-forward torques $\vec{\tau}_{ff}$ are calculated from an explicit dynamical model. However, for specific, repeatable, tasks, we can rely on specific torques to provide low trajectory tracking errors. An example of such is the CMP framework \cite{Denisa2016}, which associates the desired kinematic behavior with the corresponding joint torques for a specific task parameter. These corresponding joint torques are \emph{learned} for the specific case, and not generally applicable.
%in some cases it might be extremely difficult to obtain the relevant model. In such cases, as explained in \cite{Denisa2016}, we can rely on specific torques to provide low trajectory tracking errors for specific situations. An example of applying specific torques for specific situations is

\subsection{Compliant Movement Primitives}\label{sec:singleCMP}
For the sake of completeness we provide a brief description of CMPs. For details see \cite{Denisa2016}.

A CMP cobines desired joint motion trajectories (joint positions $\vec{q}_d(t)$) and corresponding joint torque signals $\vec{\tau}_f(t)$
\begin{equation}
\vec{h}(t) =[\vec{q}_d(t), \vec{\tau}_f(t)].
\end{equation}
Joint positions for all degrees-of-freedom (DOF) are obtained from demonstration, for example through imitation, while joint torques are recorded from a stiff execution. Note that task-specific torques are gained by substracting the known robot's $\vec{f}_{dynamic}(\vec{q}, \dot{\vec{q}}, \ddot{\vec{q}})$ from the actual measured torques $\vec{\tau}_m$ at robots joints
\begin{equation} \label{eq:tau_m}
\vec{\tau}_f = \vec{\tau}_m - \vec{f}_{dynamic}(\vec{q}, \dot{\vec{q}}, \ddot{\vec{q}}).
\end{equation}
A short discussion on this is in Section \ref{sec:discussion}.

Joint positions are encoded as dynamic movement primitives (DMPs) \cite{Ijspeert2013}, consisting of a linear part and a nonlinear forcing term composed of a combination of weighted kernel functions.  The corresponding torques are encoded as only a combination of radial basis functions (RBFs). We here omit the equations for lack of space. However, both are dependent on the same phase signal. The weights of the DMP forcing term and the weights of the RBF (in short) define their signals. We again refer the reader to \cite{Denisa2016} for details.

CMPs provide the reference trajectories and torque profiles in \emph{joint} space. However, the relation between the robots in a bimanual task is in task space and kinematic transformations are needed. Bimanual kinematics is presented in the next section. Note that when the robots are redundant for the task and offer more than one solution, we need to make sure that the robot is in the correct, desired posture. A discussion on redundancy is provided in Section \ref{sec:discussion}.

%\subsection{Feed-forward Torques for Bimanual Settings}
%The CMP control method from Section \ref{sec:singleCMP} was designed for joint-space control of the robot. It is thus not applicable for a bimanual setting, where the control has to be implemented in the task space -- otherwise it is impossible to maintain the bimanual task. To achieve the desired behavior, defined in \ref{sec:statement}, we have to modify the feed-forward torques to include thre parts:
%
%XXX - final feed-forward torque equation - made of three parts.
%
%: 1) reduce the trajectory tracking error when there are no perturbations. This is implemented through the learned CMP torques, calculated from task-space trajectories as described in Section \ref{sec:final_ctrl}. 2) maintain the bimanual task, i.\,e., the relative posture of the robots. This is implemented through the symmetric bimanual torque controller, described in Section \ref{sec:symmetric_bimanual}. Finally, we want to 3) reduce the "resistance" effect of the symmetric bimanual torque controller through virtual force translation, described in Section \ref{sec:force_translation}.

\section{SYMMETRIC ROBOT CONTROL}\label{sec:Bimanual}
This section shows the basic kinematics of a bimanual system. For a more complete description we refer the reader to \cite{Chiacchio98}.
	\begin{figure}[b]
	\centering
	\includegraphics[width = 0.4\textwidth]{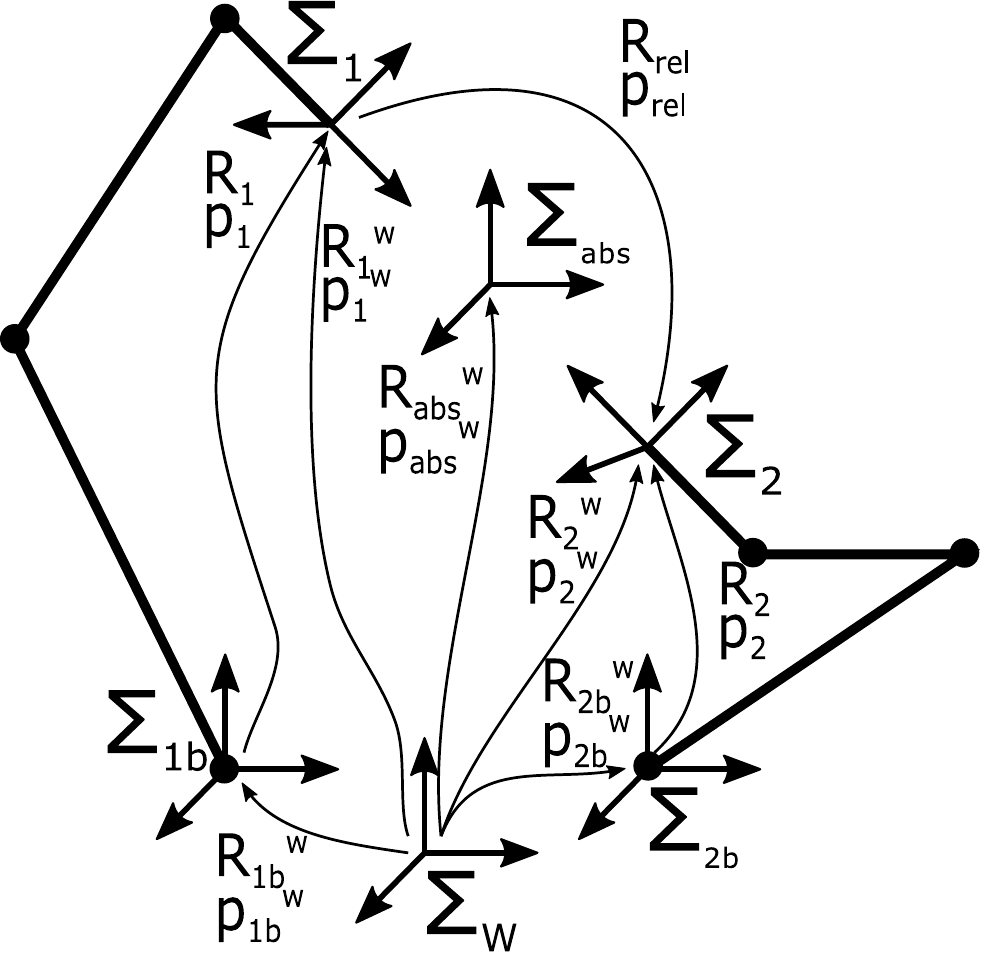}
	\caption{Schematic of a bimanual system's kinematics.}
	\label{fig:bi-man_kine}
	\end{figure}

\subsection{Kinematic Control of a Bimanual System}
The basic kinematic schematic of a bimanual system with relevant coordinate frames is shown in Fig. \ref{fig:bi-man_kine}. The coordinate frames are labelled as $\Sigma_{x}$, where the subscript $x$ stands for one of the following options: $W$ -- world, $i$ -- $i$ -th robot tool-center-point (TCP), $ib$ -- $i$-th robot base, $\mathrm{abs}$ -- absolute.
%
%	\begin{equation} \nonumber
%	\begin{matrix}
%	W, && \mathrm{world} \\
%	i, && i\mathrm{-th\ robot\ TCP}\\
%	ib, && i\mathrm{-th\ robot\ base}\\
%	abs, && \mathrm{absolute}.\\
%	\end{matrix}
%	\end{equation}	
Note that $\Sigma_{W}$ is the inertial coordinate frame of the system. The relation between these coordinate frames is given in the form of $X_Y^Z$, where $X$ stands for either the position vector $\vec{p}$, or the rotation matrix $\textbf{R}$.
%	\begin{equation}
%	 X =
%	 \left\{
%	 \begin{matrix}
%	 \vec{p}, && \mathrm{position\ vector} \\
%	 \textbf{R}, && \mathrm{rotation\ matrix,} \\
%	 \end{matrix}
%	 \right.
%	\end{equation}
The subscript $_Y$ is the self index, and the superscript $^Z$ is the index of the coordinate frame in which $X$ is described in. The robots parameters that are described in its own coordinate frame have no superscript.
	
The two most common approaches to bimanual control are 1) asymmetric and 2) symmetric control. Asymmetric control controls each robot end-effector (or TCP) independently of the other, with synchronized trajectories \cite{Gams14a}. For control in a 6 DOF bimanual task, each of the robots has to control its end-effector in 6 DOF.

Symmetric control, on the other hand, splits the desired movement so that each robot does half of the movement \cite{Chiacchio98}. In this case, the task description is separated into absolute and relative motion. The absolute coordinates (6 DOF) describe the position and orientation of a common coordinate frame of the robots in reference to the inertial (world) coordinate frame. However, the relative coordinates (6 DOF) describe the position and orientation of one robot TCP relative to the other. This is described by
	\begin{equation} \label{eq:abs_pos}
	\vec{p}_{abs}^{\,w} = \frac{1}{2}(\vec{p}_1^{\,w} + \vec{p}_2^{\,w})
	\end{equation}	
	\begin{equation} \label{eq:abs_rot}
	\textbf{R}_{abs}^{\,w} = \textbf{R}_{1}^{\,w}\textbf{R}_{\vec{k}_{12}^{1}}^{1}\left(\frac{\theta_{12}}{2}\right)
	\end{equation}	
	\begin{equation} \label{eq:rel_pos}
	\vec{p}_{rel}^{\,w} = \vec{p}_2^{\,w} - \vec{p}_1^{\,w}
	\end{equation}		
	\begin{equation} \label{eq:rel_rot}
	\textbf{R}_{r}^{1} = \textbf{R}_{2}^{1}
	\end{equation}
	In (\ref{eq:abs_rot}), $\vec{k}_{12}^{1}$ and $\theta_{12}$ are, respectively, the unit vector and the angle that realize the rotation described by $\textbf{R}_{2}^{1}$.
	
	To achieve relative motion independent of the absolute motion we use	
	\begin{equation} \label{eq:rel_pos_abs}
	\vec{p}_{rel}^{\,w} = \textbf{R}_{abs}^{\,w}\,\vec{p}_{rel}^{\,abs}
	\end{equation}
By taking the time derivatives of (\ref{eq:abs_pos}) -- (\ref{eq:rel_rot}) we gain	
	\begin{equation} \label{eq:abs_vel}
	\dot{\vec{p}}_{abs}^{\,w} = \frac{1}{2}(\dot{\vec{p}}_1^{\,w} + \dot{\vec{p}}_2^{\,w})
	\end{equation}	
	\begin{equation} \label{eq:abs_w}
	\vec{\omega}_{abs}^{\,w} = \frac{1}{2}(\vec{\omega}_1^{\,w} + \vec{\omega}_2^{\,w})
	\end{equation}	
	\begin{equation} \label{eq:rel_vel}
	\dot{\vec{p}}_{rel}^{\,w} = \dot{\vec{p}}_2^{\,w} - \dot{\vec{p}}_1^{\,w}
	\end{equation}		
	\begin{equation} \label{eq:rel_w}
	\vec{\omega}_{rel}^{\,w} = \vec{\omega}_2^{\,w} - \vec{\omega}_1^{\,w}
	\end{equation}
Thus we can implement differential kinematics, which can be applied for inverse kinematics calculation. For single robots we have
	\begin{equation} \label{eq:diff_kin}
	\begin{matrix}
	\begin{bmatrix}
	\dot{\vec{p}}_{i}^{\,w} \\ \vec{\omega}_{i}^{\,w}
	\end{bmatrix} = \textbf{J}_{i}^{w}\left(\vec{q_i}\right) \dot{\vec{q_i}}, & & i = 1, 2
	\end{matrix},
	\end{equation}
where $\dot{\vec{p}}_{i}^{\,w}$ and $\vec{\omega}_{i}^{\,w}$ are the vectors of position and angular velocity of the robot end-effector, respectively. $\dot{\vec{q_i}}$ is the vector of angular velocities in joint space.

For the coupled bimanual system we combine (\ref{eq:abs_vel}) and (\ref{eq:abs_w}) with (\ref{eq:diff_kin}). We get
	\begin{equation} \label{eq:abs_diff_kin}
	\begin{bmatrix}
	\dot{\vec{p}}_{abs}^{\,w} \\ \vec{\omega}_{abs}^{\,w}
	\end{bmatrix} = \textbf{J}_{abs}^{w}\left(\vec{q_1}, \vec{q_2}\right)
	\begin{bmatrix}
	\dot{\vec{q}}_1 \\ \dot{\vec{q}}_2
	\end{bmatrix},
	\end{equation}
	\begin{equation} \label{eq:j_abs}
	\textbf{J}_{abs}^{w} = \begin{bmatrix}
	\frac{1}{2}\textbf{J}_{1}^{w} \ \frac{1}{2}\textbf{J}_{2}^{w}
	\end{bmatrix}.
	\end{equation}
	Also, by combining (\ref{eq:rel_vel}) and (\ref{eq:rel_w}) with (\ref{eq:diff_kin}), we can derive
	\begin{equation} \label{eq:rel_diff_kin}
	\begin{bmatrix}
	\dot{\vec{p}}_{rel}^{\,w} \\ \vec{\omega}_{rel}^{\,w}
	\end{bmatrix} = \textbf{J}_{rel}^{w}\left(\vec{q_1}, \vec{q_2}\right)
	\begin{bmatrix}
	\dot{\vec{q}}_1 \\ \dot{\vec{q}}_2
	\end{bmatrix},
	\end{equation}
	\begin{equation} \label{eq:j_rel}
	\textbf{J}_{rel}^{w} = \begin{bmatrix}
	-\textbf{J}_{1}^{w} \ \textbf{J}_{2}^{w}
	\end{bmatrix}.
	\end{equation}
	Merging (\ref{eq:abs_diff_kin}) and (\ref{eq:rel_diff_kin}) results in:
	\begin{equation} \label{eq:bi-man_diff_kin}
	\begin{bmatrix}
	\dot{\vec{p}}_{abs}^{\,w} \\ \vec{\omega}_{abs}^{\,w} \\ \dot{\vec{p}}_{rel}^{\,w} \\ \vec{\omega}_{rel}^{\,w}
	\end{bmatrix} = \textbf{J}^{w}\left(\vec{q_1}, \vec{q_2}\right)
	\begin{bmatrix}
	\dot{\vec{q}}_1 \\ \dot{\vec{q}}_2
	\end{bmatrix}
	\end{equation}
	where
	\begin{equation}
	\textbf{J}^{w} = \begin{bmatrix}
	\textbf{J}_{abs}^{w} \\ \textbf{J}_{rel}^{w}
	\end{bmatrix}.
	\end{equation}
	Thus we can iteratively calculate the inverse kinematics using
	\begin{equation} \label{eq:kinematics}
	\dot{\vec{q}} = \textbf{J}^{\dag}\left(\vec{v}_d + \textbf{K}\vec{e}\right).% +% \left(\textbf{I} - \textbf{J}^{\dag} \textbf{J}\right)\dot{\vec{q}}_{0}.
	\end{equation}
	In case of redundancy, secondary tasks can be added through the null-space of the $\textbf{J}$. In (\ref{eq:kinematics}) $
	\dot{\vec{q}} =
	\begin{bmatrix}
	\dot{\vec{q}}_1^{\,T} & \dot{\vec{q}}_2^{\,T}
	\end{bmatrix}^{T}
	$ is the vector of angular velocities, $
	\vec{e} =
	\begin{bmatrix}
	\vec{e}_{abs}^{\,T} & \vec{e}_{rel}^{\,T}
	\end{bmatrix}^{T}
	$ is the vector of task space errors, $
	\vec{v}_{d} = \begin{bmatrix}
	\vec{v}_{absd}^{\,T} & \vec{v}_{reld}^{\,T}
	\end{bmatrix}^{T}
	$ is the vector of desired task space velocities, $\textbf{K}$ is a $12\times12$ diagonal gain matrix, $\textbf{I}$ is a $12\times12$ identity matrix, $\textbf{J}$ is the previously described Jacobian matrix and $\dot{\vec{q}}_{0}$ is a vector of desired joint space velocities of lower priority used in case of a redundant system. The $\dagger$ sign is used to annotate the Moore-Penrose pseudo-inverse. The errors and desired velocities are calculated by
	\begin{equation} \label{eq:abs_error}
	\vec{e}_{abs} =
	\begin{bmatrix}
	\vec{p}_{absd}^{\,w} - \vec{p}_{abs}^{\,w} \\
	\begin{split}
	\frac{1}{2}\left(\textbf{S}\left(\vec{n}_{abs}^{\,w}\right)\vec{n}_{absd}^{\,w} \right.   + \textbf{S} & \left(\vec{s}_{abs}^{\,w}\right)\vec{s}_{absd}^{\,w} + \\
	& \left. + \textbf{S}\left(\vec{a}_{abs}^{\,w}\right)\vec{a}_{absd}^{\,w}\right)
	\end{split}
	\end{bmatrix}
	\end{equation}		
	\begin{equation} \label{eq:rel_error}
	\vec{e}_{rel} =
	\begin{bmatrix}
	\textbf{R}_{abs}^{w}\vec{p}_{reld}^{\,abs} - \vec{p}_{r}^{\,w} \\
	\begin{split}
	\frac{1}{2}\textbf{R}_{1}^{w}\left(\textbf{S}\left(\vec{n}_{rel}^{\,1}\right)\vec{n}_{reld}^{\,1} + \textbf{S} \right. & \left(\vec{s}_{rel}^{\,1}\right)\vec{s}_{reld}^{\,1} + \\  & \left.+ \textbf{S}\left(\vec{a}_{rel}^{\,1}\right)\vec{a}_{reld}^{\,1}\right)
	\end{split}
	\end{bmatrix}
	\end{equation}	
	\begin{equation}
	\vec{v}_{absd} = \begin{bmatrix}
	\dot{\vec{p}}_{absd}^{\,w} \\ \vec{\omega}_{absd}^{\,w}
	\end{bmatrix}
	\end{equation}	
	\begin{equation}
	\vec{v}_{reld} = \begin{bmatrix}
	\textbf{R}_{abs}^{w}\dot{\vec{p}}_{reld}^{\,abs} + \textbf{S}\left(\vec{\omega}_{abs}^{\,w}\right)\textbf{R}_{abs}^{w}\vec{p}_{reld}^{\,abs} \\ \vec{\omega}_{reld}^{1}
	\end{bmatrix}
	\end{equation}
	where the subscript suffix $d$ stands for desired, $\textbf{S}(\cdot)$ is the skew-symmetric operator and $\vec{n}_{i}^{\,j}, \vec{s}_{i}^{\,j}, \ \vec{a}_{i}^{\,j}$ are, respectively, the first, second and third column of a rotation matrix. i.e. $\textbf{R}_{i}^{j} = \begin{bmatrix}\vec{n}_{i}^{\,j} \ \vec{s}_{i}^{\,j} \ \vec{a}_{i}^{\,j}\end{bmatrix}$.

\subsection{Symmetric Bimanual Torque Controller}\label{sec:symmetric_bimanual}
Now that we have defined the relevant kinematic variables, we can use these to implement a symmetric bimanual torque controller. The derivation of this controller is thoroughly described in \cite{Chiacchio98}.

Instead of calculating the desired joint positions to control the behavior of the robot, we calculate the joint torques. The controller is defined by
\begin{equation}\label{eq:torque_ctrler}
  \vec{\tau}_{\mathrm{biman}} = \textbf{J}^T \left(\textbf{K}_{\mathrm{task}}\left(\vec{x}_d - \vec{x}\right) + \textbf{D}_{\mathrm{task}}( \dot{\vec{x}}_d - \dot{\vec{x}}) \right)
\end{equation}
Just as in the kinematic case, $\textbf{K}_{\mathrm{task}}$ and $\textbf{D}_{\mathrm{task}}$ are $12\times12$ diagonal gain matrices for stiffness and damping, respectively (6 DOF for the absolute and 6 for the relative task), with gains $k_i$ or $d_i,~i=1, 2,...,12$, respectively, on the diagonals. A low $k_i$ will result in compliant behavior for task DOF $i$, which also means that trajectory tracking in task DOF $i$ results in high errors.

The controller increases joint torques based on the error in task space. Since the gains are decoupled for separate DOFs, in case of low gains for the absolute DOFs, $K_{\mathrm{abs}} << K_{\mathrm{rel}}$ the robot will be compliant in absolute space, but stiff in relative space. However, it will also not be able to track the desired trajectories in the absolute space. As described in Section \ref{sec:CMPs}, trajectory tracking is ensured through the torque part of CMPs.

The drawback of this controller is that symmetric control changes the torques of both manipulators. This means that pushing on one robot will result in additional torques in both robots, appearing in order to to neutralize the perturbation. A push on a robot, for example from an unintentional collision, will thus result in less compliant behavior of the bimanual system. To increase the compliancy of the bimanual system, we introduce also a virtual force translation.

\subsection{Virtual Force Translation}\label{sec:force_translation}
Through measuring of joint torques we can estimate the end-effector force using the virtual work theorem. In a general case it states
\begin{equation}\label{eq:virtual_work}
\vec{\tau} = \textbf{J}^{T}\vec{f}_{e}
\end{equation}
In case of a perturbation on one robot, we can thus estimate the
end-effector force of one manipulator $\vec{f}_{1e}$ using
(\ref{eq:virtual_work}). We can now apply the same end-effector force
through the joint torques to the other robot. Thus we have
\begin{equation}\label{eq:virtual_force_equality}
\vec{f}_{1e} = \vec{f}_{2e}.
\end{equation}
From (\ref{eq:virtual_work}) we get
\begin{equation}\label{eq:virtual_work_inverse}
\vec{f}_{e} = \left(\textbf{J}^{\dagger}\right)^{T}\vec{\tau}
\end{equation}
and substituting (\ref{eq:virtual_work_inverse}) into
(\ref{eq:virtual_force_equality}) we get
\begin{equation}\label{eq:virtal_force_equality_torque}
\left(\textbf{J}_{1}^{\dagger}\right)^{T}\vec{\tau}_{1} =
\left(\textbf{J}_{2}^{\dagger}\right)^{T}\vec{\tau}_{2}
\end{equation}
from which we can derive
\begin{eqnarray}
\vec{\tau}_{1e} &=&
\textbf{J}_1^{T}\left(\textbf{J}_2^{\dag}\right)^{T}\vec{\tau}_{2e} \label{eq:tau1}, \\
\vec{\tau}_{2e} &=&
\textbf{J}_2^{T}\left(\textbf{J}_1^{\dag}\right)^{T}\vec{\tau}_{1e} \label{eq:tau2}.
\end{eqnarray}
It should be noted that only the virtual torques caused by the perturbation
should be translated to the other robot. These are calculated by
\begin{equation}
\Delta\vec{\tau}_{i} = \vec{\tau}_{i\_{\mathrm{expected}}} -
\vec{\tau}_{i\_{\mathrm{measured}}}, ~  i = 1, 2.
\end{equation}
Substituting this into (\ref{eq:tau1}) and (\ref{eq:tau2}) we finally get
\begin{equation} \label{eq:vft_final}
\vec{\tau}_{\mathrm{vft}} =  \begin{bmatrix}
	\vec{\tau}_{\mathrm{vfc},1} \\ \vec{\tau}_{\mathrm{vfc},2}
	\end{bmatrix} =
\begin{bmatrix}
\textbf{J}_{1}^{T}\left(\textbf{J}_{2}^{\dagger}\right)^{T}\Delta\vec{\tau}_{2}
\\
\textbf{J}_{2}^{T}\left(\textbf{J}_{1}^{\dagger}\right)^{T}\Delta\vec{\tau}_{1}
\end{bmatrix}.
\end{equation}
We can see in (\ref{eq:vft_final}) that a perturbation on robot 2 $\Delta\vec{\tau_2}$ results in changed control torques in robot 1.

\subsection{CMP-based Bimanual control}\label{sec:final_ctrl}
We now have all the necessary components to construct a controller that satisfies our problem statement. Relying on (\ref{eq:kuka_ctrl2}), repeated here
\begin{equation} \nonumber
	\label{eq:kuka_ctrl3}
	\vec{\tau}_u = \textbf{K}_q(\vec{q}_d - \vec{q}\,) + \textbf{D}_q(\dot{\vec{q}}_d - \dot{\vec{q}\,}) - \vec{f}_{dynamic}(\vec{q}, \dot{\vec{q}}, \ddot{\vec{q}}\,) + \vec{\tau}_{ff},
	\end{equation}
we use the feed-forward $\vec{\tau}_{ff}$ and reduce $\textbf{K}_q$ and $\textbf{D}_q$ to achieve compliance, but still preserve accurate trajectory tracking for specific, learned tasks.

Feed-forward torques $\vec{\tau}_{ff}$ are now composed of three components
\begin{eqnarray}
  \vec{\tau}_{ff} &=& \begin{bmatrix}
	\vec{\tau}_{ff,1} \\ \vec{\tau}_{ff,2}
	\end{bmatrix} = \vec{\tau}_{\mathrm{rec}} + \vec{\tau}_{\mathrm{biman}} - \vec{\tau}_{\mathrm{vft}}. \label{eq:final_ctrl}
%   &=& \vec{\tau}_{\mathrm{rec}} + \textbf{J}^{T}(\textbf{K}_{task}(\vec{x}_d - \vec{x}) + \textbf{D}_{task}(\dot{\vec{x}}_d - \dot{\vec{x}})) - \vec{\tau}_{fvc}
\end{eqnarray}

The pre-recorded or learned task torque	$\vec{\tau}_{\mathrm{rec}}$ ensures trajectory tracking. It is the direct output of the CMP. However, the reference joint trajectories are calculated from the task-space trajectories using (\ref{eq:kinematics}). Again note that the inverse kinematics solution needs to match the posture of the robot during the demonstration\footnote{Even though we used a 14 DOF system in our experiments, we only used 12 DOF, locking the rotation of the 3rd axis on both robots. Thus our system was not redundant for the task.}. This can be achieved using a secondary task
	\begin{equation} \label{eq:kinematics2}
	\dot{\vec{q}} = \textbf{J}^{\dag}\left(\vec{v}_d + \textbf{K}\vec{e}\right) + \left(\textbf{I} - \textbf{J}^{\dag} \textbf{J}\right)\textbf{K}_s(\vec{q}_{\mathrm{demo}} - \vec{q}_{\mathrm{act}}),
	\end{equation}
where $\mathrm{demo}$ stands for demonstrated and ${\mathrm{act}}$ for actual joint positions, with $\textbf{K}_s$ a diagonal gain matrix.

The bimanual symmetric controller $\vec{\tau}_{\mathrm{biman}}$ maintains the bimanual task. It is given with (\ref{eq:torque_ctrler}). To ensure that (\ref{eq:torque_ctrler}) does not act against (\ref{eq:kuka_ctrl2}), we set $\textbf{K}_q << \textbf{K}_{\mathrm{task}}$. In case of a redundant system we can set $\textbf{K}_{\mathrm{task,~abs}}<<\textbf{K}_{\mathrm{task,~rel}}$ and put the posture of the robot in the secondary task with
\begin{equation}\label{eq:torque_ctrler2}
\begin{split}
  \vec{\tau}_{\mathrm{biman}} = \textbf{J}^T \left(\textbf{K}_{\mathrm{task}}\left(\vec{x}_d - \vec{x}\right) + \textbf{D}_{\mathrm{task}}( \dot{\vec{x}}_d + \dot{\vec{x}}) \right)
    \\ +\left(\textbf{I} - \left(\textbf{J}^{\dag}\right)^T \textbf{J}^T\right)\textbf{K}_{s1}\textbf{K}_q(\vec{q}_{\mathrm{demo}} - \vec{q}_{\mathrm{act}})
   \end{split}
\end{equation}
with $\textbf{K}_{s1}$ a diagonal gain matrix.

The virtual force translation $\vec{\tau}_{\mathrm{vft}}$ reduces the necessary feedback reaction of (\ref{eq:torque_ctrler}) and thus increases compliance of the bimanual system. It is given with (\ref{eq:vft_final}). Due to different postures of the robot, the copied virtual force only partially accounts for the perturbation, but still have a considerable effect on the compliance of the complete bimanual system.

In the ideal example, when there is no perturbation and no sensor noise, we can achieve the same behavior by only using the pre-recorded or learned task torque	$\vec{\tau}_{\mathrm{rec}}$.

%\section{GENERALIZATION}\label{sec:generalization}
%In this section we provide just a brief description of generalization of CMPs using GPR, due to lack of space. We refer the reader to \cite{rasmussen2006} for details on GPR and to \cite{Denisa2016} for details on CMP generalization.
%
%Since CMPs provide trajectories and torque profiles only for specific solutions, their applicability would remain limited to exactly the same conditions if it were not for generalization. As described in \cite{Denisa2016}, statistical generalization using GPR can be effectively applied to single-robot tasks. Given a database of CMPs and associated task parameters (e.\,g., a new CMP is recorded for every new load the robot is carrying), we can then use statistical generalization to calculate CMPs for the loads between the recorded ones. However, the behavior of the robot and the query must transition continuously between the task parameters. In this paper we apply the exact same approach for bimanual tasks. This can be defined as a function
%\begin{equation}
%\mathbf{F}_{h^N}: c \longmapsto [\vec{h}], \label{eq:generalization}
%\end{equation}
%which uses a database of $N$ CMPs (the weights of $N$ DMP forcing terms and the weights of $N$ RBFs for each robot DOF) to define a new CMP for a different task parameter $c$ (often referred to as query point), given by a new set of weights. The calculation of the new weights is performed using GPR \cite{rasmussen2006}.

\section{RESULTS}\label{sec:results}
Our experimental setup consisted of two Kuka LWR-4 7 DOF robots, as shown in Fig. \ref{fig:intro}. In our experiments we locked the rotation of the 3rd axis on both robots. Thus our system was not redundant for the task. The system was controlled from Matlab at an average at 500Hz. The robot was controlled in joint-stiffness mode, with the stiffness set to 25~Nm/rad for all the used joints, which is even lower than what was used for single robots \cite{Denisa2016}.

The task of the bimanual system was to perform a bimanual trajectory while conforming to the task description as given in Section \ref{sec:intro}. The robots were each carrying a 2.5~kg load. To emphasize the maintaining of the relative task, they were not physically coupled through holding a common object. In this experiment we only controlled the relative position of the system, while not controlling the orientation. See also Section \ref{sec:discussion}.

\begin{figure}[t]
    \vspace{2mm}
	\centering
	\includegraphics[width = 0.45\textwidth]{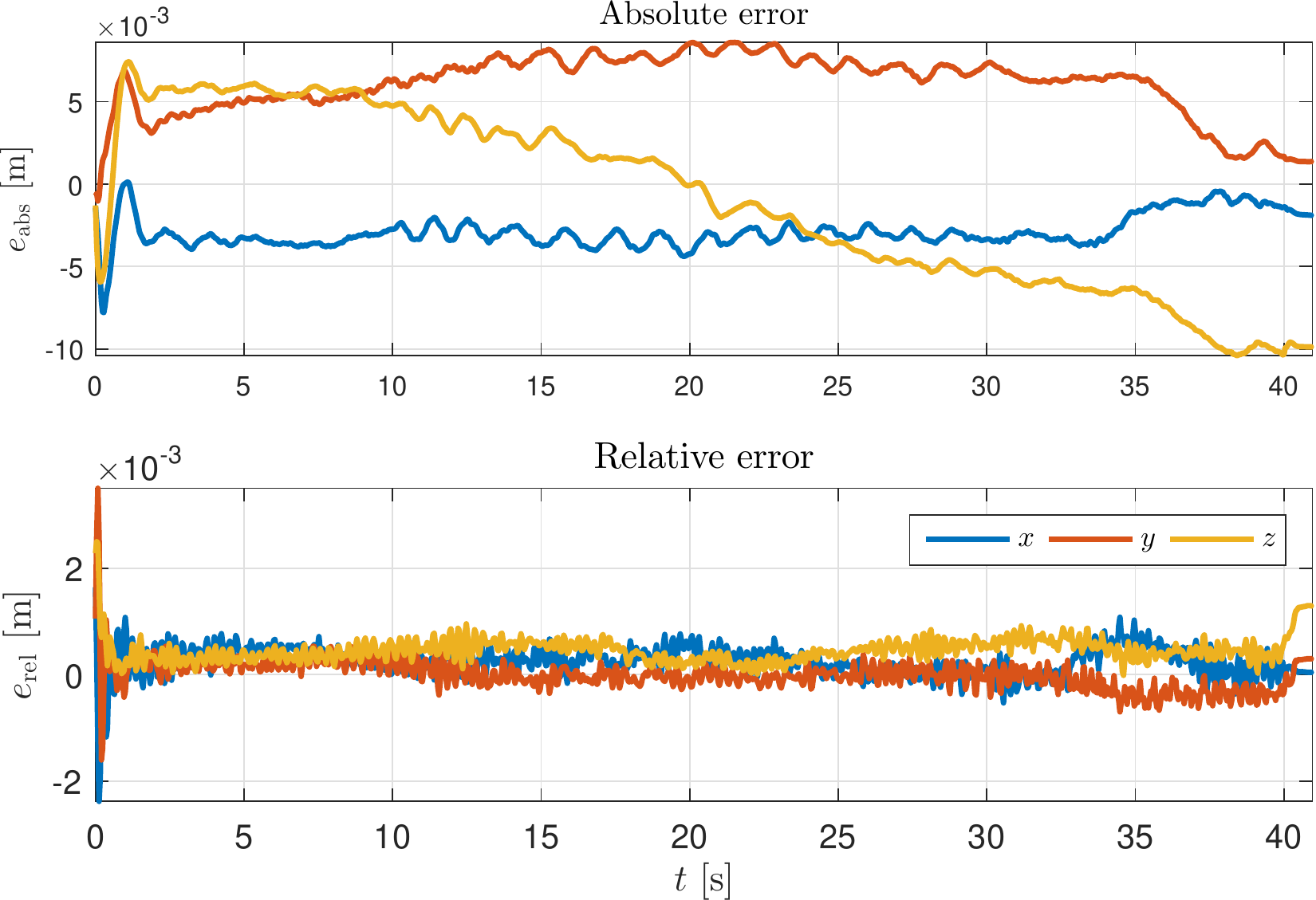}
	\caption{Absolute (top) and relative errors when there is no perturbation to the system.}
	\label{fig:orig_no_pert}
	%\vspace{-6mm}
	\end{figure}

\begin{figure}[t]
    %\vspace{2mm}
	\centering
	\includegraphics[width = 0.45\textwidth]{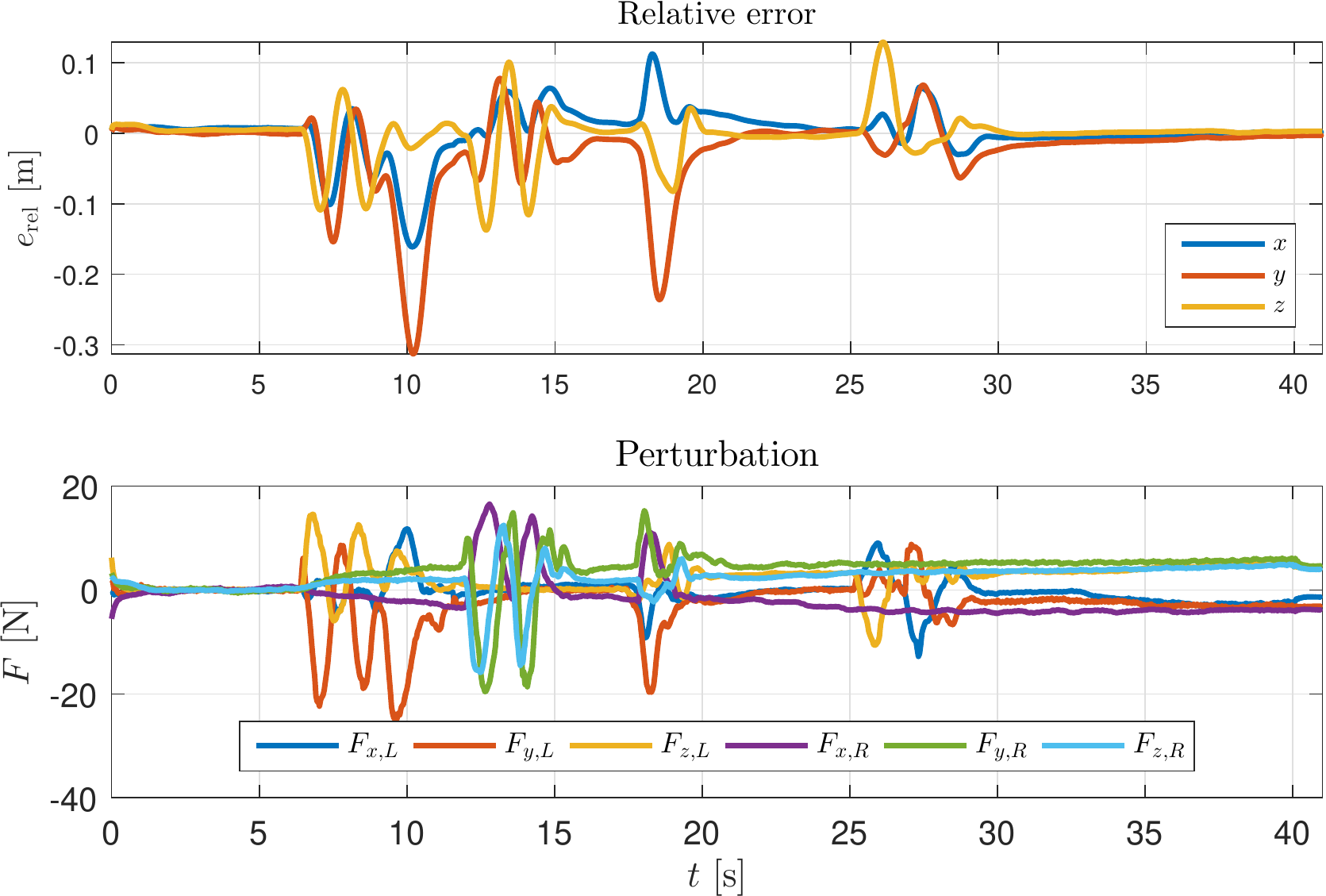}
	\caption{Relative error (top) and end-effector perturbation (calculated from measured joint torques) when using just $\tau_{ff} = \tau_{\mathrm{rec}}$.}
	\label{fig:just_cmp}
	\vspace{-2mm}
	\end{figure}

When there are no perturbations, the system follows both the absolute and the relative tasks, even if we use only $\tau_{ff} = \tau_{\mathrm{rec}}$. This is shown In Fig. \ref{fig:orig_no_pert}.
However, if perturbations occur, the system does not maintain the relative task, see Fig. \ref{fig:just_cmp}

\begin{figure}[t]
    \vspace{-2mm}
	\centering
	\includegraphics[width = 0.45\textwidth]{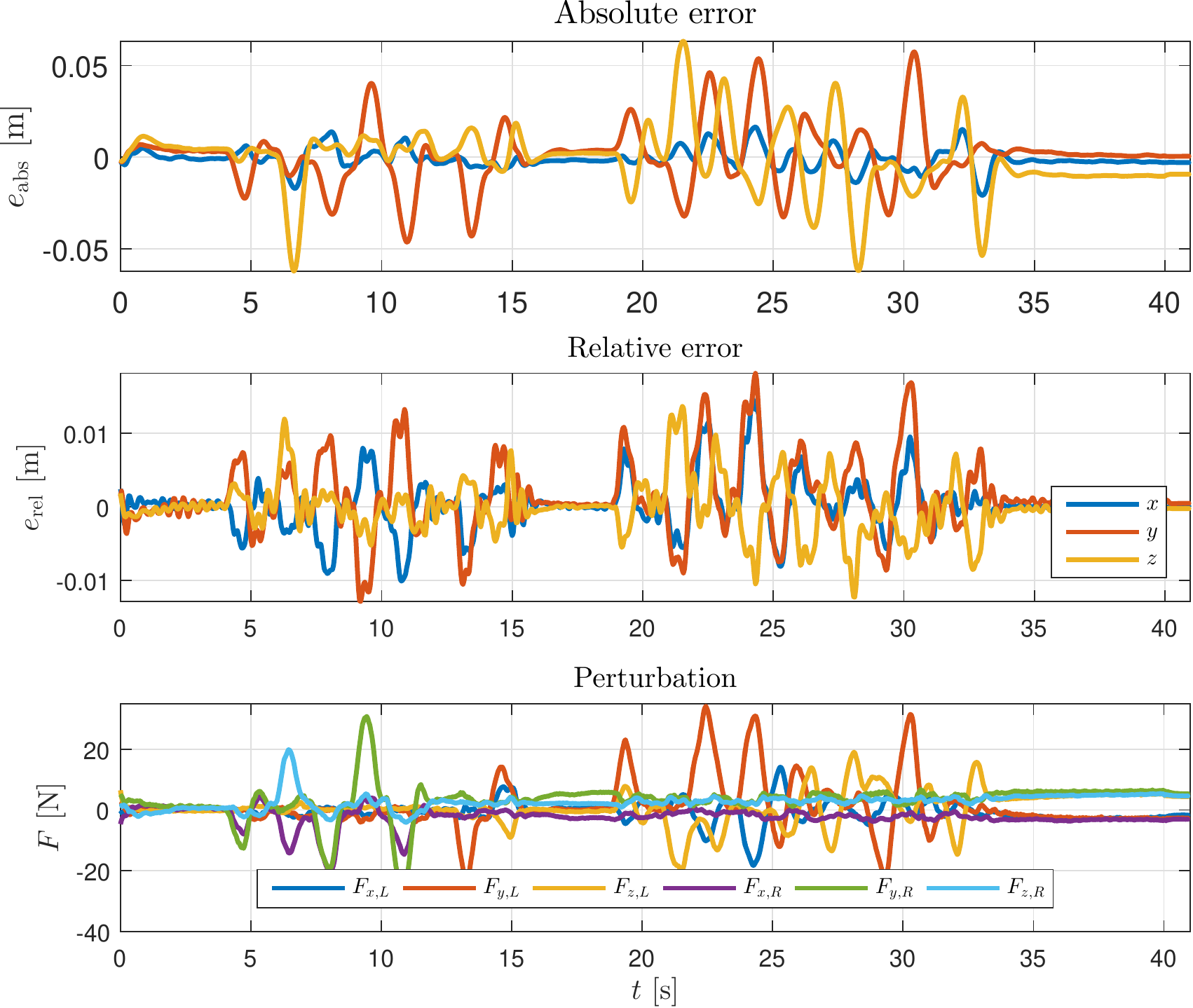}
	\caption{Absolute error (top), relative error (middle) and end-effector perturbation (calculated from measured joint torques) when using $\vec{\tau}_{ff} = \vec{\tau}_{\mathrm{rec}} + \vec{\tau}_{\mathrm{biman}}$. }
	\label{fig:NS_symmetric}
	\vspace{-2mm}
	\end{figure}

Including the symmetric bimanual torque controller $\vec{\tau}_{ff} = \vec{\tau}_{\mathrm{rec}} + \vec{\tau}_{\mathrm{biman}}$ will result in maintaining the relative task, but in reduced compliance, as shown in Fig. \ref{fig:NS_symmetric}
\begin{figure}[htb!]
    %\vspace{-2mm}
	\centering
	\includegraphics[width = 0.45\textwidth]{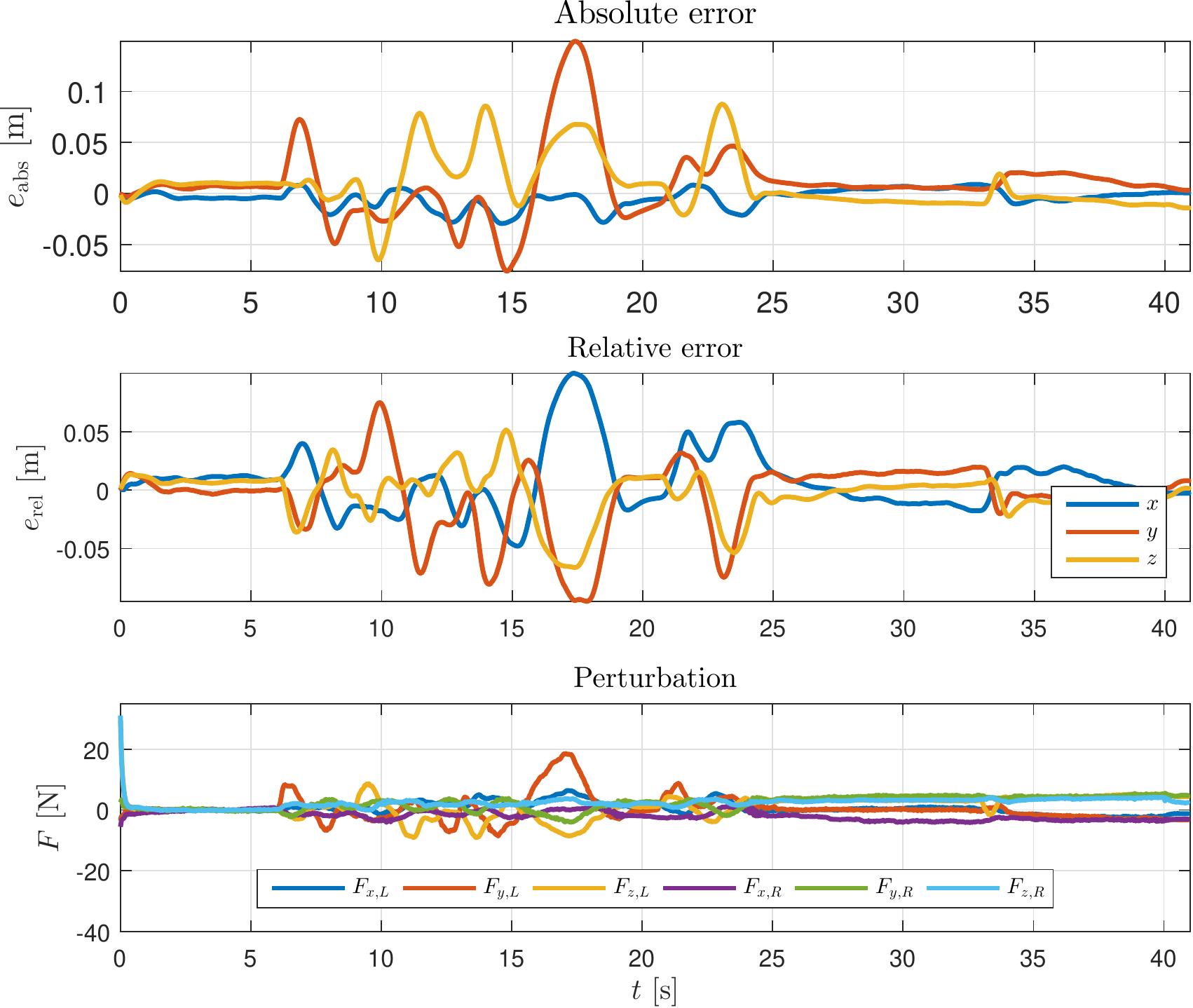}
	\caption{Absolute error (top), relative error (middle) and end-effector perturbation (calculated from measured joint torques) when using $\vec{\tau}_{ff} = \vec{\tau}_{\mathrm{rec}} - \vec{\tau}_{\mathrm{vfc}}$.}
	\label{fig:force_copy}
	\vspace{-2mm}
	\end{figure}
\begin{figure}[htb!]
    \vspace{-2mm}
	\centering
	\includegraphics[width = 0.45\textwidth]{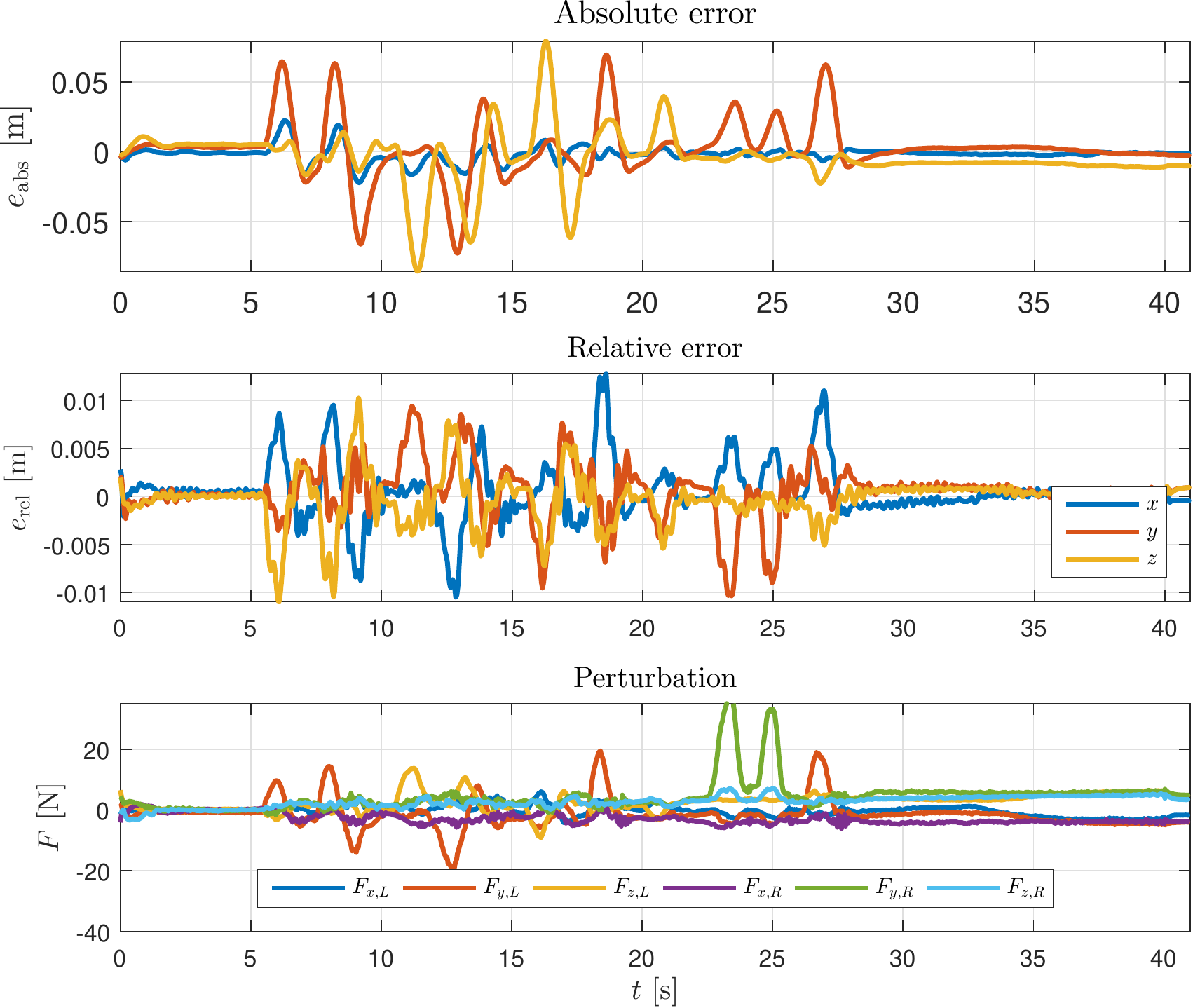}
	\caption{Absolute error (top), relative error (middle) and end-effector perturbation (calculated from measured joint torques) when using the complete controller, given by (\ref{eq:final_ctrl}).}
	\label{fig:complete}
	\vspace{-2mm}
	\end{figure}
\begin{figure}[htb!]
    %\vspace{-2mm}
	\centering
	\includegraphics[width = 0.45\textwidth]{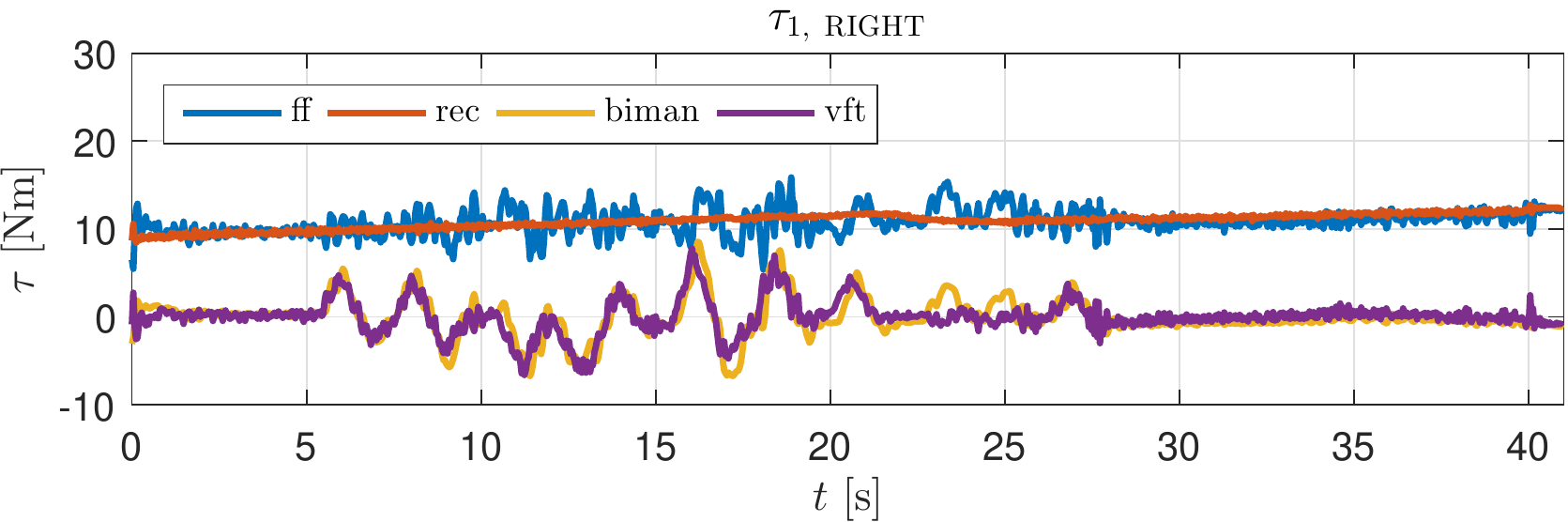}
	\caption{The feed-forward torque on the first joint of the right robot, and separate components when using the complete controller, given by (\ref{eq:final_ctrl}). See the bottom plot of Fig. \ref{fig:complete} for the perturbation.}
	\label{fig:torques_complete}
	\vspace{-2mm}
	\end{figure}

\begin{figure*}[t]
\includegraphics[width=0.15\textwidth]{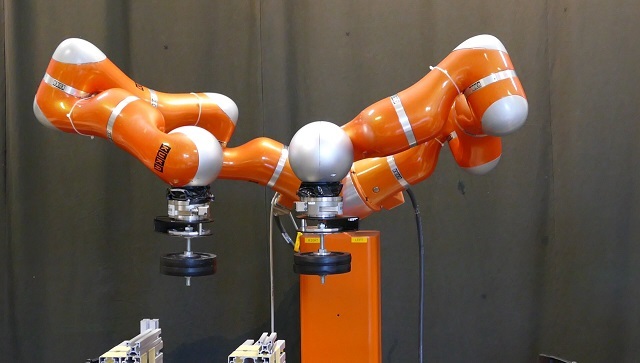}  \hfill
\includegraphics[width=0.15\textwidth]{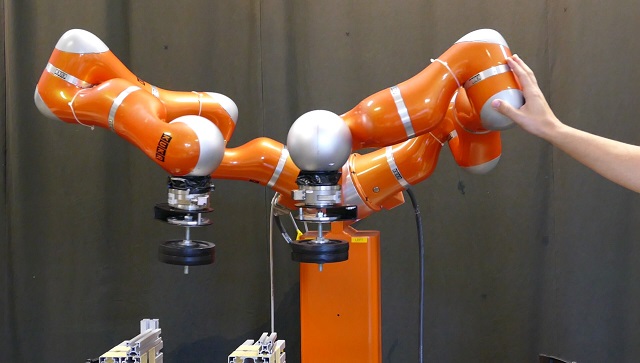}  \hfill
\includegraphics[width=0.15\textwidth]{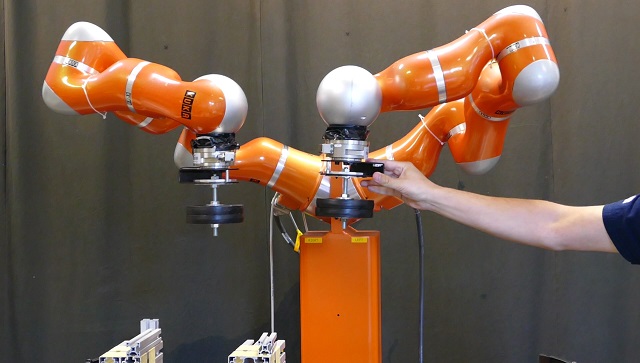}  \hfill
\includegraphics[width=0.15\textwidth]{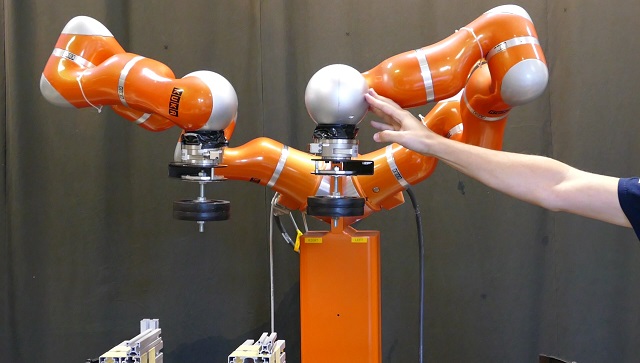}  \hfill
\includegraphics[width=0.15\textwidth]{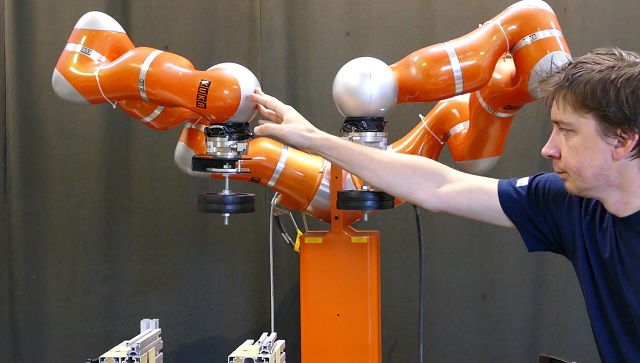}  \hfill
\includegraphics[width=0.15\textwidth]{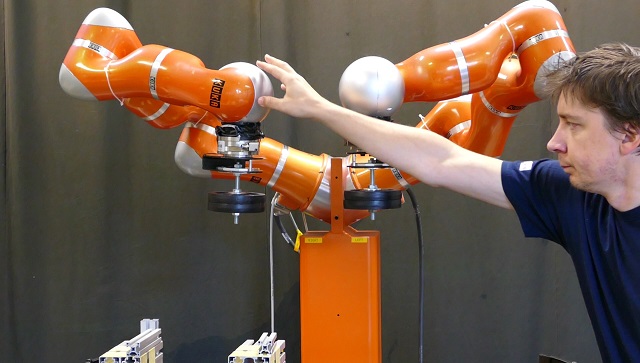}  \hfill \\[1mm]%
\includegraphics[width=0.15\textwidth]{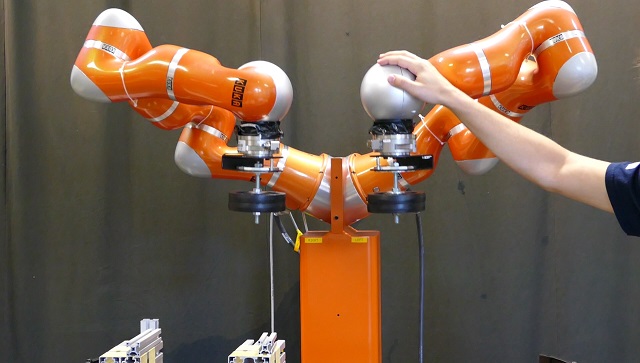}  \hfill
\includegraphics[width=0.15\textwidth]{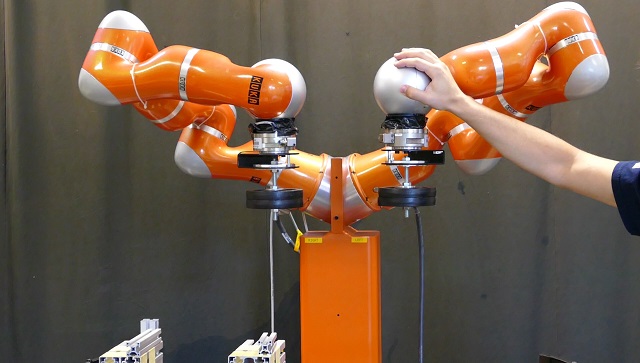}  \hfill
\includegraphics[width=0.15\textwidth]{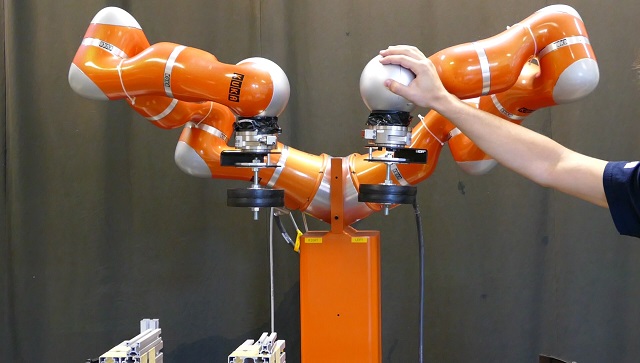}  \hfill
\includegraphics[width=0.15\textwidth]{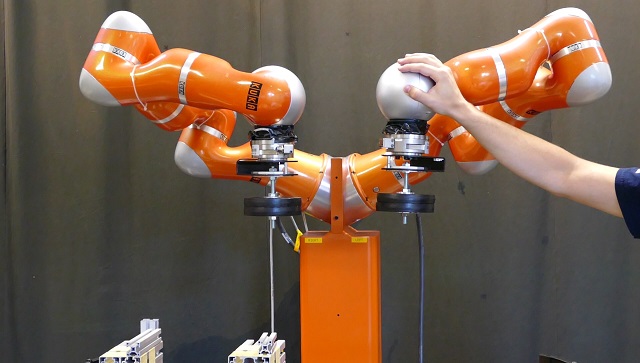}  \hfill
\includegraphics[width=0.15\textwidth]{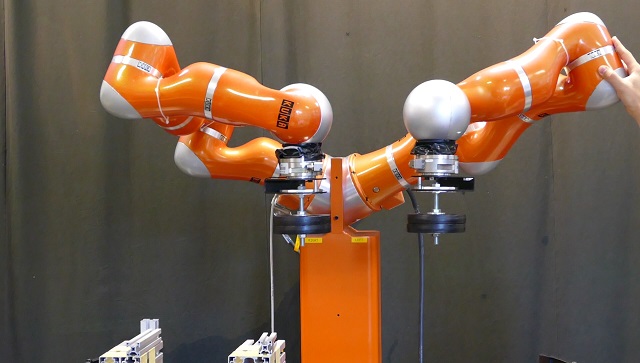}  \hfill
\includegraphics[width=0.15\textwidth]{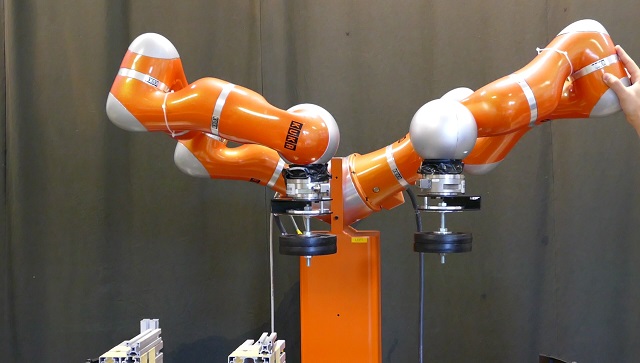}  \hfill \\\\[-1mm]%
\vspace{-6mm}
\caption{Still images of the bimanual system performing a compliant bimanual task, with a person perturbing the motion.} \label{fig:sequence}
\vspace{-4mm}
\end{figure*}

Excluding the symmetric bimanual torque controller but including the virtual force translation, $\vec{\tau}_{ff} = \vec{\tau}_{\mathrm{rec}} - \vec{\tau}_{\mathrm{vft}}$, as shown in Fig. \ref{fig:force_copy}, results in higher compliance, shown in higher absolute error for a smaller perturbation. On the other hand, the relative error is significantly higher than when using the symmetric bimanual torque controller.

The results in Table \ref{tb:Rel_Error} show that the complete system was compliant and had the best error rejection for the relative task.
	\begin{table*}[bth!]
	\centering
	\caption{Absolute and relative errors at maximum perturbation amplitude.}
	\label{tb:Rel_Error}
	\begin{tabular}{|c|c|c|c|c|} \hline
		& $\vec{\tau}_{\mathrm{ff}} = \vec{\tau}_{\mathrm{rec}}$ & $\vec{\tau}_{\mathrm{ff}} = \vec{\tau}_{\mathrm{rec}} + \vec{\tau}_{\mathrm{biman}}$ & $\vec{\tau}_{\mathrm{ff}} = \vec{\tau}_{\mathrm{rec}} - \vec{\tau}_{\mathrm{vft}}$ & Entire controller \\ \hline
\textbf{Absolute error [m]} & $/$ & 0.0563 & 0.1516 & 0.0688 \\ \hline		
\textbf{Relative error [m]} &  0.1629 & 0.0246 & 0.1433 & 0.0118 \\ \hline
\textbf{Perturbation [N]} & 27.6563 & 35.9163 & 21.4106 & 20.4994 \\ \hline
	\end{tabular}
	\end{table*}

The complete system, given by (\ref{eq:final_ctrl}) is compliant in the absolute task, but maintains low errors in the relative task despite the high forces. Thus it fully complies with the given problem statement in Section \ref{sec:intro}. The results are shown in Fig. \ref{fig:complete}. The difference between including or excluding the virtual force translation is seen also at seconds 23 -- 25 in the bottom plot of Fig. \ref{fig:complete}, where the right robot did not include it. Thus, a twice higher perturbation resulted in a much lower absolute error, meaning that the system was less compliant, when $\vec{\tau}_{\mathrm{vfc}}$ was not included.

In Fig. \ref{fig:torques_complete} we can see the complete $\tau_{1,\mathrm{RIGHT}}$ and the contributions of separate components for the first joint of the right robot. We can see that $\tau_{\mathrm{vfc}}$ and $\tau_{\mathrm{biman}}$ are similar. This means that when perturbing the left robot, we are not fighting $\vec{\tau}_{\mathrm{biman}}$ of the right robot, because it it only has to account for a much smaller relative error, which remains due to different postures of the robots that make force-vector copy inaccurate. The results for other joints are less similar (not presented). We can also see that when there is no perturbation, the contribution of the bimanual symmetric torque controller and of the virtual force translation is 0. The plot also shows that $\tau_{\mathrm{rec}}$ is the actual learned feed-forward torque, while the other two react to perturbations.

Figure \ref{fig:sequence} shows a series of still photos showing the bimanual execution and the physical interaction of a person.

\section{DISCUSSION and CONCLUSION}\label{sec:discussion}
We have shown how we can extend the CMP framework to include bimanual task execution. Even though the robots were not physically coupled through holding a rigid object, we showed that the system is compliant while maintaining the relative behavior of the robots. While the presented approach is applicable for a specific, pre-learned task, generalization has the potential to extend it for a wider region. Generalization of the bimanual CMP approach is part of our future work. Several other topics of the approach offer different possibilities.

CMPs in essence provide only one, or through generalization, a limited set of solutions. The question on whether it is better to invest more into the derivation of the dynamical model or simply learn the torque for a small set of task parameters is very valid. Typically, the dynamics of the robot can be calculated. Therefore CMPs only cover the dynamics of the task. If the task can easily be modeled, there is no advantage in using CMPs. The example with weights shown in the accompanying video could easily have been modelled. However, imagine a task that is very difficult to model, such as lifting a soft object or even a person. There, the model is difficult to obtain and there approaches such as CMPs are useful. Furthermore, if this object or person between the robots must not be pressed on from either side (squeezed) under any condition, bimanual CMP control as proposed in this paper is the correct tool.

Since the system relies on task space coordinates, but CMPs provide joint space trajectories, the mapping between the two needs to allow only one solution. In our experiments we made sure of that by locking one DOF of each robot, thus gaining a 12 DOF system for a 12 DOF task. However, when the robots are redundant for the task, kinematic mapping offers numerous solutions. Learning of torques for all solutions is not viable, as there could literally be infinite. As shown in Section \ref{sec:final_ctrl}, maintaining the posture of the robot can be achieved through the secondary task. However, CMPs act more as an enabler for softer, compliant collisions and as such as a safety mechanism. Once perturbed, a dynamical motion (unlike quite static motion in the presented experiments) will not return to the original trajectory, because the torques will not align with kinematic variables anymore until the end of the motion. As such, we only need to make sure that the robot maintains the demonstrated posture during the motion and basically ignore the motion after the perturbation, if it occurs, because it will not return to the demonstrated motion.

Performing the experiments on the real robot needs foremost a fast control loop. The robot controller, provided by the manufacturer given by (\ref{eq:kuka_ctrl2}) runs at a very high frequency. If we move that to a system with a lower frequency, the operation is not smooth and may become unstable. This is specially evident for the maintaining of the relative orientations of the robot, which were not maintained in the experiments.

In the future we wish to, besides expanding the framework with generalization, also reduce the effect of the bimanual symmetric controller through modifying the virtual force translation to account for the different postures of the two robots.

\addtolength{\textheight}{-12cm}   % This command serves to balance the column lengths

%%%%%%%%%%%%%%%%%%%%%%%%%%%%%%%%%%%%%%%%%%%%%%%%%%%%%%%%%%%%%%%%%%%%%%%%%%%%%%%%

%%%%%%%%%%%%%%%%%%%%%%%%%%%%%%%%%%%%%%%%%%%%%%%%%%%%%%%%%%%%%%%%%%%%%%%%%%%%%%%%

%%%%%%%%%%%%%%%%%%%%%%%%%%%%%%%%%%%%%%%%%%%%%%%%%%%%%%%%%%%%%%%%%%%%%%%%%%%%%%%%%
%\section*{APPENDIX}
%\section*{ACKNOWLEDGMENT}
%
%The preferred spelling of the word ÒacknowledgmentÓ in America is without an ÒeÓ after the ÒgÓ. Avoid the stilted expression, ÒOne of us (R. B. G.) thanks . . .Ó  Instead, try ÒR. B. G. thanksÓ. Put sponsor acknowledgments in the unnumbered footnote on the first page.

%%%%%%%%%%%%%%%%%%%%%%%%%%%%%%%%%%%%%%%%%%%%%%%%%%%%%%%%%%%%%%%%%%%%%%%%%%%%%%%%
\bibliographystyle{ieeetr}
\bibliography{IEEEabrv,mybibfile}

\end{document}